\documentclass[letterpaper]{article}
\PassOptionsToPackage{table}{xcolor}
\usepackage[preprint]{aaai2027}
\usepackage[hyphens]{url}
\usepackage{graphicx}
\usepackage{natbib}
\usepackage{caption}
\usepackage{amsmath}
\usepackage{amssymb}
\usepackage{booktabs}
\usepackage{cleveref}
\usepackage{xspace}
\makeatletter
\DeclareRobustCommand\onedot{\futurelet\@let@token\@onedot}
\def\@onedot{\ifx\@let@token.\else.\null\fi\xspace}

\makeatother

\newcommand{\keywords}[1]{}

\definecolor{fusion}{RGB}{210,210,210}
\definecolor{guidance}{RGB}{190,210,240}
\definecolor{backbone}{RGB}{200,230,200}
\definecolor{negative}{RGB}{255,217,102}

\begin{document}

\title{STAND: Semantic Anchoring Constraint with Dual-Granularity Disambiguation for Remote Sensing Image Change Captioning} 

\author{
Yanpei Gong\textsuperscript{\rm 1},
Beichen Zhang\textsuperscript{\rm 1}\corresponding,
Hao Wang\textsuperscript{\rm 1},
Xuhang Fu\textsuperscript{\rm 1},\\
Zhaobo Qi\textsuperscript{\rm 1},
Xinyan Liu\textsuperscript{\rm 1},
Yuanrong Xu\textsuperscript{\rm 1},
Weigang Zhang\textsuperscript{\rm 1}
}
\affiliations{
\textsuperscript{\rm 1}Harbin Institute of Technology, Weihai, China\\
2023211640@stu.hit.edu.cn, beiczhang@hit.edu.cn
}

\maketitle

\begin{abstract}
Remote sensing image change captioning (RSICC) aims to describe the difference between two remote sensing images. 
While recent methods have explored video modeling, they largely overlook the inherent ambiguities in viewpoint, scale, and prior knowledge, lacking effective constraints on the encoder.
In this paper, we present STAND, a Semantic Anchoring Constraint with Dual-Granularity Disambiguation for RSICC, to progressively resolve these ambiguities.
Specifically, to establish a reliable feature foundation, we first introduce an interpretable constraint to regularize temporal representations.
Operating on these purified features, a dual-granularity disambiguation module resolves spatial uncertainties by coupling macro-level global context aggregation for viewpoint confusion with micro-level frequency-refocused attention for small-object scale enhancement.
Ultimately, to translate these visually disambiguated features into precise text, a semantic concept anchoring module leverages language categorical priors to tackle knowledge ambiguity during decoding.
Extensive experiments verify the superiority of STAND and its effectiveness in addressing viewpoint, scale, and knowledge ambiguities. 
The code will be publicly released on GitHub.

\end{abstract}

\section{Introduction}
\label{sec:intro}

Remote sensing image change captioning (RSICC) aims to describe the difference between two bi-temporal images. 
With increasing demand in urban monitoring and environmental assessment, this task has gained growing attention in both the remote sensing (RS) and vision-language communities~\cite{shi2024multi,liu2022remote}.

\begin{figure}[t]
  \centering
   \includegraphics[width=0.92\linewidth]{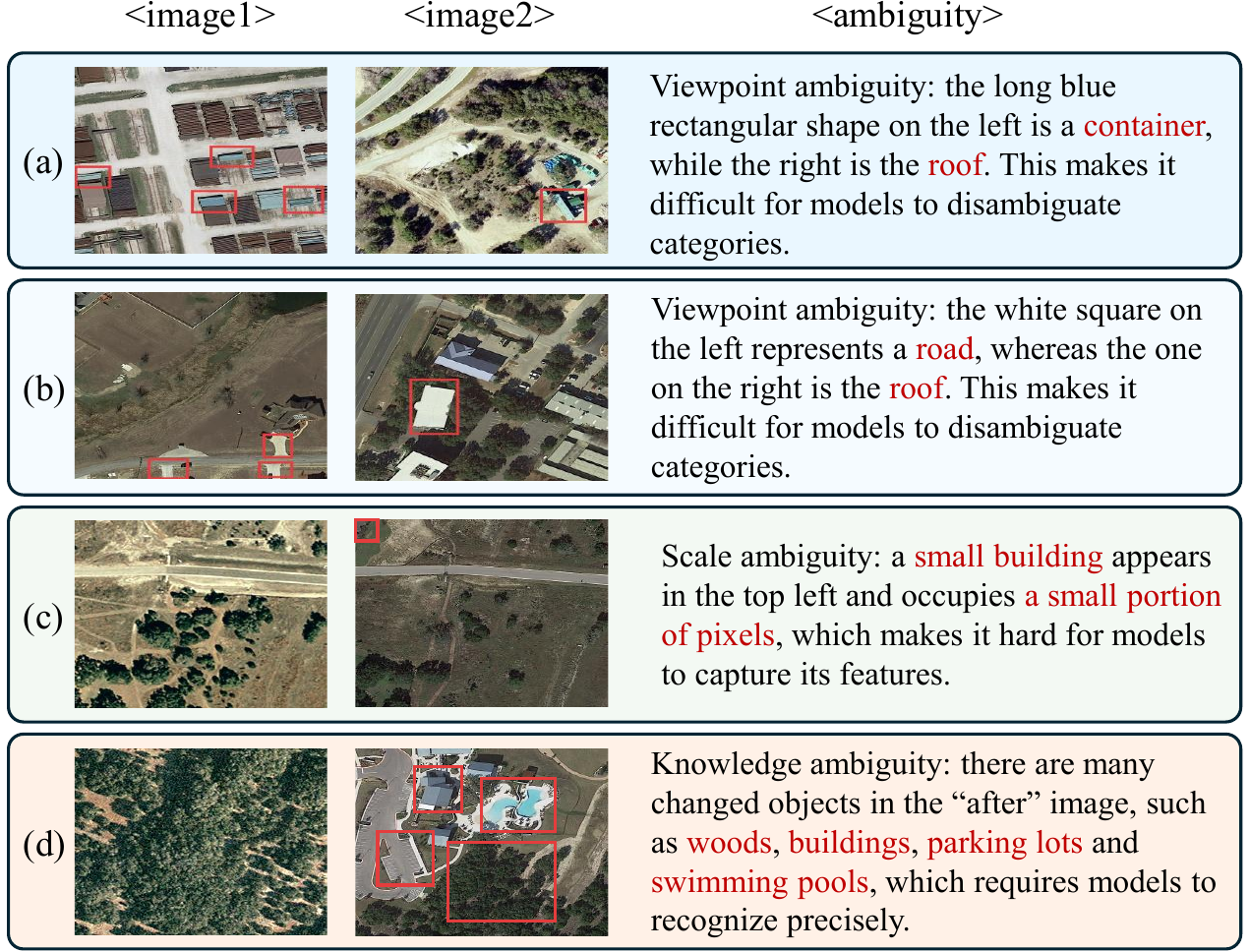}

   \caption{Typical examples of ambiguities in the remote sensing images. We attribute these ambiguities to three main factors: viewpoint, scale, and knowledge.}
   \label{fig1}
\end{figure}

Unlike natural image change captioning, RSICC is more challenging due to the inherent ambiguities within RS images. 
We attribute these ambiguities to three factors: viewpoint, scale, and knowledge.
(1) Viewpoint ambiguity (\cref{fig1} (a) (b)): since distinct objects may have highly similar appearances under a top-down and remote viewpoint, RS images are often confusing.
(2) Scale ambiguity (\cref{fig1} (c)): a 256*256 image often corresponds to a region with hundreds of square meters~\cite{liu2022remote,shi2024multi}, while changed objects occupy only a small portion of pixels making the changes difficult to distinguish and recognize from the background.
(3) Knowledge ambiguity (\cref{fig1} (d)): In contrast to natural image settings~\cite{park2019robust}, RSICC requires precise entity-level identification, making it heavily dependent on domain-specific prior knowledge, especially when multiple entities appear. 

To generate accurate change descriptions, most previous studies~\cite{chang2023changes,att1zhou2024single,liu2022remote} adopt a static image encoder for bi-temporal modeling.
Nevertheless, image encoders mainly focus on local feature extraction while neglecting long-range context~\cite{yang2024remote,zhu2025change3d}.
Building upon such encoders, difference-centric methods~\cite{att4chen2024multi,chang2023changes,liu2022remote,liu2023progressive} assume that visual discrepancies directly correspond to semantic changes.
However, this assumption often fails in RS due to the aforementioned ambiguity, where distinct objects may have a similar appearance from top-down viewpoint.
Therefore, we argue that resolving such an issue requires understanding the overall scene rather than difference features alone.
Recently, mask-based methods~\cite{li2025cd4c,liu2024mv} attempted to highlight changed regions by hard filtering.
Yet, this operation ignores contextual information that is crucial for disambiguating look-alike objects, often amplifying the ambiguity.
More recently, video-based models utilizing learnable perception frames were proposed to extract temporal features~\cite{zhu2025change3d}. While they capture spatio-temporal dynamics, they naively overlook the inherent ambiguities common in remote sensing and lack effective constraints on the encoder, failing to systematically decouple these domain-specific challenges.

Based on these observations, we argue that high-quality change captioning requires progressively resolving these ambiguities.
Accordingly, we propose STAND, a \textbf{S}eman\textbf{T}ic \textbf{A}nchoring Co\textbf{N}straint with \textbf{D}ual-Granularity Disambiguation for RSICC. 
First, to capture temporal change dynamics, STAND models the before-mask-after triplet as a short video clip and encodes it using a video encoder. 
To further regularize the temporal transition, an Interpretable Transition Constraint (ITC) module encourages consistent relations among before, change, and after representations.
Building upon these representations, we design a Dual-Granularity Target Disambiguation (DGTD) module to address scale and viewpoint ambiguities. Specifically, a macro-level Context-Aware Viewpoint Disambiguation (CAVD) module incorporates global semantics to distinguish visually similar entities, while a micro-level Frequency-Refocused Complementary Attention (FRCA) enhances small-object responses to mitigate scale ambiguity.
Finally, to resolve knowledge ambiguity, a Semantic Concept Anchoring (SCA) module leverages language categorical priors to guide the decoding process.

Overall, STAND addresses the three ambiguities through a progressive pipeline and outperforms state-of-the-art methods on extensive RS datasets.

The main contributions are summarized as follows:
\begin{itemize}
	\item We propose STAND, a spatio-temporal disambiguation framework for RSICC that jointly addresses viewpoint, scale, and knowledge ambiguities. Its transition constraint regularizes representations to capture change dynamics.
	
	\item We develop a progressive disambiguation strategy that couples dual-granularity spatial reasoning with semantic concept anchoring. DGTD combines global context with small-object refinement, while SCA grounds visual changes in structured language priors.
	
	\item We conduct extensive experiments together with ambiguity-specific evaluations, demonstrating consistent improvements under small-scale changes, viewpoint ambiguity, and knowledge ambiguity.
\end{itemize}

\section{Related Work}
\label{sec:relatedwork}

%-------------------------------------------------------------------------
\subsection{Change Captioning} \label{ssec:cc}

Change captioning is an essential task in the field of vision language processing~\cite{wang2026aifindartifactawareinterpretingfinegrained, tu2024context, zhu2025change3d}. Depending on the image domain, existing research in this area can be broadly divided into two main domains based on image type: natural images and remote sensing images~\cite{chang2023changes}.

In the domain of natural images, early works~\cite{li2025region,tu2021semantic,hosseinzadeh2021image,lv2025revisiting,tu2023adaptive,zhong2025decider} have explored various methods for localizing and describing changes.
%A primary challenge is extracting features that represent the changes. To address this, 
A typical approach~\cite{kim2021agnostic,yue2023i3n,shi2020finding} is to isolate the change by computing the image similarity and removing the shared components. 
Another approach~\cite{tu2023SCORER} adopts contrastive learning for better robustness.
These advances in natural images have offered valuable insights, yet their applicability to other domains remains limited due to domain-specific ambiguities.

In view of this, remote sensing image change captioning (RSICC) was proposed for satellite images. 
Unlike natural scenes, remote sensing (RS) images contain more complex, dynamic landscapes, substantial visual noise and unrelated changes like illumination and seasonal effects \cite{chang2023changes,liu2025remote}.
Therefore, the goal is to describe meaningful changes while ignoring unimportant visual variations. 
Recent studies~\cite{att1zhou2024single,att2li2024inter,liu2023progressive,att4chen2024multi,chang2023changes} employ various attention mechanisms or revisit the task from a video-modeling perspective~\cite{zhu2025change3d} to enhance difference perception.
In parallel, multi-task frameworks~\cite{mutil1wang2024changeminds,shi2024multi,mutil3karaca2025robust,mutil4sun2025mask} leverage complementary supervision from related tasks to improve representation learning.
More recently, works~\cite{li2025cd4c,liu2024mv} have introduced change masks to highlight the changed areas and achieved promising results. Yet, their methods rely on explicit mask filtering, which risks disrupting contextual integrity and causes irreversible information loss.
Furthermore, most RS methods~\cite{shi2024multi,liu2022remote,tu2024context,chang2023changes} adopt Siamese encoders that process the two images separately. 
However, this parallel processing architecture focuses on local feature extraction while often neglecting crucial scene-level long-range context~\cite{zhu2025change3d,yang2024remote}.

\begin{figure*}[t]
\centering
\includegraphics[width=0.92\textwidth]{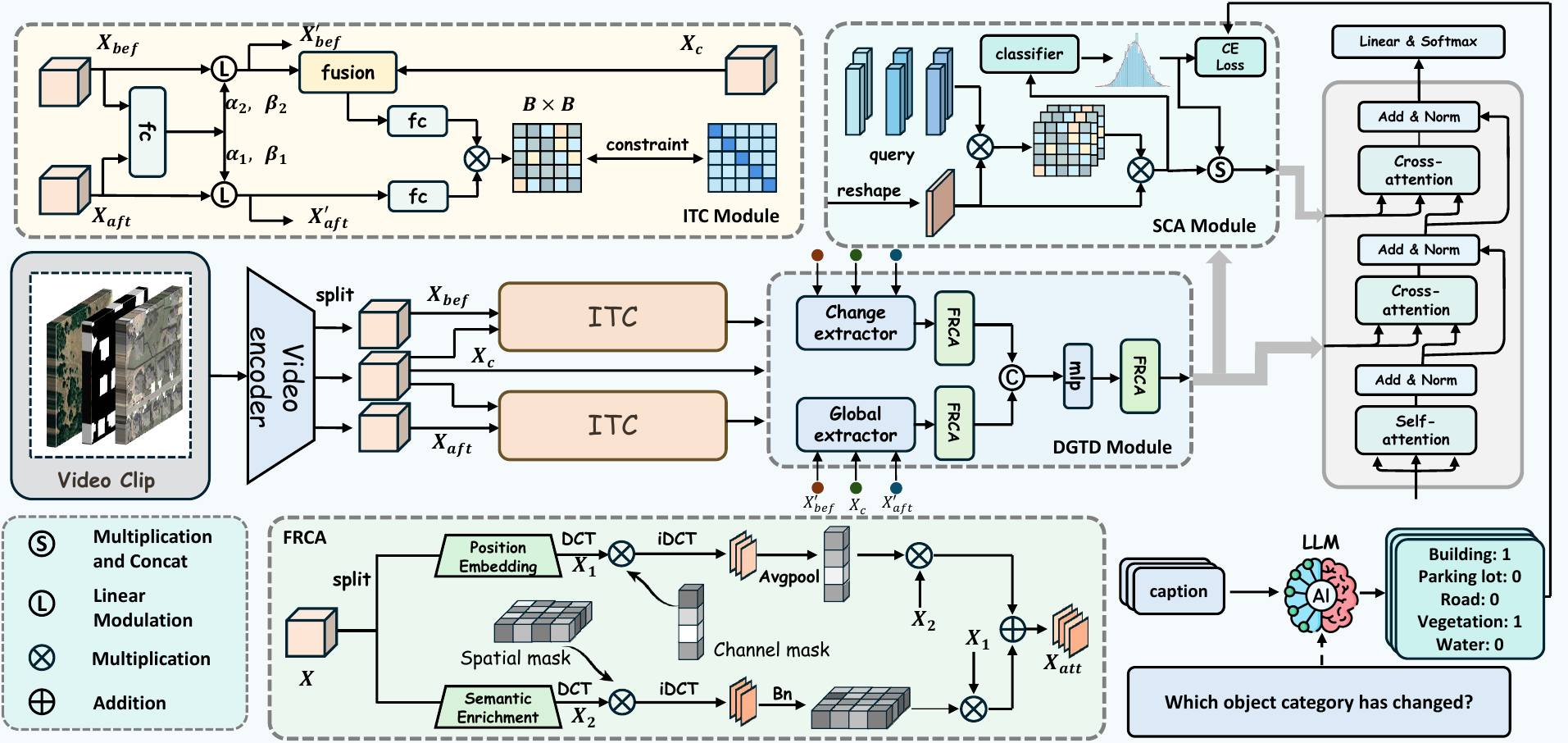} % Reduce the figure size so that it is 
  \caption{The architecture of STAND, including an interpretable transition constraint (ITC), dual-granularity target disambiguation (DGTD) module, and semantic concept anchoring (SCA) module. $B$ and fc denote batch size and fully connected layer.}
\label{fig2}
\end{figure*}

%-------------------------------------------------------------------------
\subsection{Video Understanding} \label{ssec:videou}

Video understanding provides an effective paradigm for modeling spatio-temporal dynamics \cite{tang2025video}.
By treating a sequence of frames as a unified input, video understanding models excel at capturing spatio-temporal relationships and dynamic events.
% why we adapt video understanding for change captioning
Early works~\cite{video1,video2,video3,video4} have extended 2D CNNs to 3D CNNs, achieving notable progress but suffering from high computational cost. 
To achieve a better trade-off between accuracy and efficiency, \cite{feichtenhofer2020x3d} proposed X3D to expand 2D image backbone along several network axes. 
This design enables powerful spatio-temporal feature learning while keeping computational complexity low, making it well suited for compact video-like modeling scenarios.

\section{Methodology}
\label{sec:Method}

As illustrated in~\cref{fig2}, STAND follows a two-stage pipeline.
In the first stage, a pretrained video encoder encodes the short video clip, the ITC regularizes the temporal representations, and the DGTD module resolves scale and viewpoint ambiguities.
In the second stage, the SCA module grounds the disambiguated visual representations in language priors and guides the decoder to generate change descriptions.

\subsection{Video Clip Encoding}

Formally, given a pair of bi-temporal images $I_{\text{bef}}$ and $I_{\text{aft}}$, along with the ``change mask'' $I_{\text{mask}}$, 
STAND first concatenates them to form a short video clip.  
The concatenated sequence is then fed into a video encoder and the encoder outputs are subsequently split into three feature tokens, denoted as $X_{\text{bef}}$, $X_{\text{c}}$, and $X_{\text{aft}}$. This process is defined by:
\begin{equation}
\big(X_{\text{bef}}, X_{\text{c}}, X_{\text{aft}}\big)
= \operatorname{Split}\!\left(\mathcal{F}_{\text{enc}}([I_{\text{bef}}; I_{\text{mask}}; I_{\text{aft}}])\right),
\label{eq:video_encoding}
\end{equation}
where $\mathcal{F}_{\text{enc}}(\cdot)$ is a video encoder, and $[\cdot ; \cdot]$ denotes concatenation. $\textit{I}\in \mathcal{R}^{3 \times H \times W}$, $\textit{X} \in \{\mathcal{R}^{ C_i \times \frac{H}{2^{i+1}} \times \frac{W}{2^{i+1}}}\}_{i=0}^3$. \textit{$X_i^t$} indicates the feature of the $t\text{-th}$ frame from the $i\text{-th}$ encoder layer. $I_{\text{mask}}$ is obtained from a pre-trained change detector~\cite{zhu2025change3d} for explicit change guidance.

\subsection{Interpretable Transition Constraint} \label{section:reconstruct}

After video clip encoding, we apply FiLM-style linear modulation~\cite{perez2018film} between $X_{\text{bef}}$ and $X_{\text{aft}}$, with feed-forward networks predicting element-wise scale and bias to suppress feature entanglement from 3D convolutions.

To supervise this modulation, we synthesize a pseudo-after feature from the corresponding before and change features:

\begin{equation} \label{eq:fusion}
	X^{(k)}_{\text{p-aft}} = \mathcal{F}_{\text{fus}}(X_{\text{bef}}^{(k)}, X_{c}^{(k)}),
\end{equation}
This explicitly models the transition: the composition should recover its paired after state in a sample-specific manner, rather than merely enforcing similarity between unspecified latent representations.
We use AFF~\cite{dai2021attentional} for feature fusion.
The pair $\big(X^{(k)}_{\text{aft}}, X^{(k)}_{\text{p-aft}}\big)$ is treated as a
positive sample, while all other features $X^{(j)}_{\text{p-aft}}$ ($j \neq k$) serve as negatives.
The model projects $X^{(k)}_{\text{aft}}$ and $X^{(k)}_{\text{p-aft}}$ into a shared embedding space,
denoted as $\tilde{X}^{(k)}_{\text{aft}}$ and $\tilde{X}^{(k)}_{\text{p-aft}}$ .

We adopt a bidirectional InfoNCE loss \cite{oord2018representation} to
align with the true after state within the batch, while pushing away all mismatched pairs:
\begin{equation} \label{eq:info_nce}
\begin{aligned}
    \mathcal{L}_{p2a} &= -\frac{1}{B} \sum_{k=1}^{B} \log \frac{e^{\operatorname{sim}(\tilde{X}^{(k)}_{\text{aft}}, \tilde{X}^{(k)}_{\text{p-aft}}) / \tau}}{\sum_{j=1}^{B} e^{\operatorname{sim}(\tilde{X}^{(k)}_{\text{aft}}, \tilde{X}^{(j)}_{\text{p-aft}}) / \tau}}, \\
    \mathcal{L}_{a2p} &= -\frac{1}{B} \sum_{k=1}^{B} \log \frac{e^{\operatorname{sim}(\tilde{X}^{(k)}_{\text{p-aft}}, \tilde{X}^{(k)}_{\text{aft}}) / \tau}}{\sum_{j=1}^{B} e^{\operatorname{sim}(\tilde{X}^{(k)}_{\text{p-aft}}, \tilde{X}^{(j)}_{\text{aft}}) / \tau}}, \\
    \mathcal{L}_{\text{ca}} &= \mathcal{L}_{p2a} + \mathcal{L}_{a2p},
\end{aligned}
\end{equation}
where ``sim'' is the dot-product to measure the similarity between two features. $\tau$ is a temperature hyper-parameter.

\subsection{Dual-Granularity Target Disambiguation Module}

Although ITC aligns temporal features, viewpoint and scale ambiguities remain in top-down RS images. DGTD addresses them through macro-level context aggregation and micro-level frequency attention, respectively.

\subsubsection{Micro-Level: Frequency-Refocused Complementary Attention} \label{subsec:frca}
Continuous down-sampling weakens the high-frequency details that distinguish small changed objects, allowing low-frequency background responses to dominate.
To mitigate this scale ambiguity, Frequency-Refocused Complementary Attention (FRCA) adaptively reweights spatial and channel frequency responses and combines them through complementary gating.
Given $X \in \mathcal{R}^{C \times H \times W}$, we split its channels at a ratio of $\gamma:(1-\gamma)$ and project the two parts back to $C$ channels, obtaining $X_s$ and $X_c$, respectively.
Here, $\gamma$ is a hyper-parameter governing channel allocation.

For frequency refinement, we construct a radial spatial mask $M_s \in \mathcal{R}^{1 \times H \times W}$ and a learnable channel-frequency mask $M_c \in \mathcal{R}^{C \times 1 \times 1}$.
Let $d_{x,y}$ denote the normalized radial distance from frequency coordinate $(x,y)$ to the DC component. The masks are defined as:
\begin{equation}
\begin{gathered}
d_{x,y} = \frac{1}{\sqrt{2}}
\sqrt{\left(\frac{x}{H-1}\right)^2+\left(\frac{y}{W-1}\right)^2}, \\
\textit{HighPass}_{x,y} = \sigma\!\left(\kappa(d_{x,y}-t)\right), \\
\textit{LowPass}_{x,y} = 1-\textit{HighPass}_{x,y}, \\
M_s = \textit{HighPass}+(1-\beta)\textit{LowPass},
\\
M_c = \sigma(\theta_c), \quad \theta_c \in \mathcal{R}^{C \times 1 \times 1},
\end{gathered}
\label{eq:mask}
\end{equation}
where $\sigma(\cdot)$ is the sigmoid function, $\kappa$ is a hyper-parameter controlling the filter sharpness, $t\in(0,1)$ is the cutoff threshold, and $\beta$ and $\theta_c$ are learnable.
$\theta_c$ is initialized such that $M_c$ is close to one. Within each FRCA instance, both masks are shared across samples.
We then reweight frequency responses in the spatial and channel dimensions:
\begin{equation}
\begin{gathered}
X_s' = \operatorname{iDCT}_{HW}\!\left(\operatorname{DCT}_{HW}(X_s) \odot M_s\right), \\
X_c' = \operatorname{iDCT}_{C}\!\left(\operatorname{DCT}_{C}(X_c) \odot M_c\right),
\end{gathered}
\label{eq:dct}
\end{equation}
where $\operatorname{DCT}_{HW}$ and $\operatorname{DCT}_{C}$ denote the 2D spatial DCT and 1D channel-wise DCT, respectively; $\operatorname{iDCT}_{HW}$ and $\operatorname{iDCT}_{C}$ are their inverses.

Finally, following complementary interaction~\cite{xiao2025fbrt}, the channel summary of $X_c'$ modulates $X_s$, while the spatial response of $X_s'$ modulates $X_c$:
\begin{equation}
X_{\text{att}} = X_s \odot \sigma(\operatorname{AvgPool}(X_c')) + X_c \odot \sigma(\operatorname{BN}(\operatorname{Conv}(X_s'))),
\end{equation}
where $\operatorname{BN}(\cdot)$ denotes BatchNorm and $\operatorname{Conv}$ reduces the channel dimension to 1.

\subsubsection{Macro-Level: Context-Aware Viewpoint Disambiguation}
Micro-level refinement strengthens small-object responses, but local evidence remains ambiguous for objects with similar top-down appearances.
CAVD therefore uses the change feature to retrieve scene-wide evidence from both temporal features, flattened into $HW$ tokens.

CAVD forms an ordered difference representation and a temporally balanced context representation:
\begin{equation}
\begin{gathered}
D_{\text{bef}} =
\mathcal{A}_{d,1}(X_{\text{c}},X_{\text{bef}},X_{\text{bef}}),\\
C_{\text{diff}} =
\mathcal{A}_{d,2}(D_{\text{bef}},X_{\text{aft}},X_{\text{aft}}),\\
C_t = \mathcal{A}_{g}(X_{\text{c}},X_t,X_t),
\quad t\in\{\text{bef},\text{aft}\},
\end{gathered}
\end{equation}
where the three arguments of each multi-head cross-attention operation denote query, key, and value, respectively.
$\mathcal{A}_{d,1}$ and $\mathcal{A}_{d,2}$ sequentially preserve the before-to-after order, while $\mathcal{A}_{g}$ shares parameters across the two temporal inputs.

We combine $C_{\text{bef}}$ and $C_{\text{aft}}$ through a feature-wise gate:
\begin{equation}
\begin{gathered}
G = \sigma\!\left(\phi_g([C_{\text{bef}};C_{\text{aft}}])\right),\\
C_{\text{global}} =
\phi_o\!\left([G\odot\phi_v(C_{\text{bef}});
(1-G)\odot\phi_v(C_{\text{aft}})]\right),
\end{gathered}
\end{equation}
where $\phi_g$, $\phi_v$, and $\phi_o$ are linear projections, $\phi_v$ is shared by both inputs, and $[\cdot;\cdot]$ denotes channel concatenation.

The ordered branch retains directional difference cues, while the gated branch supplies scene-wide context.
FRCA serves as the micro-level refinement operator when combining these two representations:
\begin{equation}
\begin{gathered}
C_{\text{diff}}' = \operatorname{FRCA}_{d}(C_{\text{diff}}),\quad
C_{\text{global}}' = \operatorname{FRCA}_{g}(C_{\text{global}}),\\
C_{\text{truth}} = \operatorname{FRCA}_{o}\!\left(
\phi_f([C_{\text{diff}}';C_{\text{global}}'])\right),
\end{gathered}
\end{equation}
where $\phi_f$ is a linear projection, and each subscript denotes an independently parameterized FRCA instance.

\subsection{Semantic Concept Anchoring Module} \label{sec:query}
Although DGTD resolves spatial ambiguity, visual features alone may not reliably identify the changed entities.
SCA therefore converts $C_{\text{truth}}$ into category-indexed visual tokens using category-change labels obtained only from the training set; label construction is described in the experimental setup.
Let $\tilde{C}_{\text{truth}}\in\mathcal{R}^{HW\times C}$ be the flattened visual feature.
For category $i$, a learnable query set $Q_i\in\mathcal{R}^{n\times C}$ aggregates category-specific evidence:
\begin{equation}
\begin{gathered}
A_i = \operatorname{Softmax}_{HW}\!\left(
\frac{Q_i\tilde{C}_{\text{truth}}^{\top}}{\tau\sqrt{C}}\right),\\
z_i = \operatorname{LN}\!\left(
\operatorname{AvgPool}_{n}(A_i\tilde{C}_{\text{truth}})\right),
\end{gathered}
\end{equation}
where $A_i\in\mathcal{R}^{n\times HW}$ is normalized over spatial locations, $n$ is the number of queries per category, and $\tau$ is a learnable temperature.
$\operatorname{LN}$ and $\operatorname{AvgPool}_{n}$ denote LayerNorm and averaging over the $n$ queries, respectively.

To estimate whether category $i$ has changed, $z_i$ is concatenated with its learnable category embedding $e_i$ and classified into change or no-change.
The resulting probability $\mathbf{p}_i\in\mathcal{R}^{2}$ weights the category token:
\begin{equation}
\begin{gathered}
\mathbf{p}_i = \operatorname{Softmax}\!\left(f_{\text{cls}}([z_i;e_i])\right), \\
\bar{z}_i = \operatorname{sg}(p_{i,1})\,z_i,\:
C_{\text{object}} = \operatorname{Stack}(\bar{z}_1,\ldots,\bar{z}_N),
\end{gathered}
\end{equation}
where $p_{i,1}$ is the change probability, $\operatorname{sg}$ denotes stop-gradient, and $C_{\text{object}}\in\mathcal{R}^{N\times C}$ contains the $N$ category tokens.

As the final step of concept-guided decoding, each decoder layer attends first to $C_{\text{truth}}$ and then to $C_{\text{object}}$:
\begin{equation}
\begin{gathered}
H_l' = \operatorname{MHCA}(\operatorname{MHSA}(H_{l-1}), C_{\text{truth}}),\\
H_l = \operatorname{MHCA}(H_l', C_{\text{object}}),
\end{gathered}
\label{eq:decoder}
\end{equation}

Here, MHSA denotes multi-head self-attention, and the two-input MHCA uses its second input as both key and value.
$H_{l-1}$ and $H_l$ denote the input and output of layer $l$, respectively.
The top-layer representation is projected to vocabulary probabilities for caption generation.

\subsection{Joint Training}
% The proposed STAND framework is trained in an end-to-end manner to jointly optimize caption generation, representation learning, and object reasoning. 
Given the ground-truth caption $(w_1^*, \dots, w_m^*)$, the decoder minimizes the negative log-likelihood:
\begin{equation}
\mathcal{L}_{\text{cap}}(\theta) = - \sum_{t=1}^{m} \log p_{\theta}(w_t^* \mid w_{<t}^*),
\label{eq:lcap}
\end{equation}
where $\theta$ denotes learnable parameters. Category-change classification uses:
\begin{equation}
\mathcal{L}_{\text{cls}} = -\frac{1}{N}\sum_{i=1}^{N}\sum_{j=1}^{2}
w_{i,j}y_{i,j}\log p_{i,j},
\label{eq:cls}
\end{equation}
where $y_{i,j}$ is the binary target for category $i$ and $w_{i,j}$ balances the classes. The total objective is:
\begin{equation}
\mathcal{L} = \mathcal{L}_{\text{cap}} + \lambda_{\text{ca}}\mathcal{L}_{\text{ca}} + \lambda_{\text{cls}}\mathcal{L}_{\text{cls}},
\label{eq:total_loss}
\end{equation}
where $\lambda_{\text{ca}}$ and $\lambda_{\text{cls}}$ are hyper-parameters.% that balance caption supervision, contrastive reconstruction, and object reasoning.

\begin{table*}[t]
	\centering
	\small
	\renewcommand{\arraystretch}{1.15}
	\begin{tabular}{l|ccccccc}
		\hline
		\rule{0pt}{12pt} \footnotesize Method & \footnotesize BLEU-1 & \footnotesize BLEU-2 &\footnotesize BLEU-3 &\footnotesize BLEU-4 & \footnotesize METEOR & \footnotesize ROUGE$_{L}$ & \footnotesize CIDEr \\
		\hline
		%DUDA~\cite{park2019robust} 
		%& 81.44 
		%& 72.22 
		%& 64.24 
		%& 57.79 
		%& 37.15 
		%& 71.04 
		%& 124.32 \\
		
		%RSICCFormer~\cite{liu2022remote} 
		%& 84.72 
		%& 76.27 
		%& 68.87 
		%& 62.77 
		%& 39.61 
		%& 74.12 
		%& 134.12 \\
		
		%PromptCC~\cite{liu2023decoupling} 
		%& \cellcolor[rgb]{ .988,  .894,  .839}85.10 
		%& \cellcolor[rgb]{ .988,  .894,  .839}77.05 
		%& \cellcolor[rgb]{ .988,  .894,  .839}70.01 
		%& \cellcolor[rgb]{ .988,  .894,  .839}64.09 
		%& \cellcolor[rgb]{ 1,  .933,  .859}39.59 
		%& \cellcolor[rgb]{ .988,  .894,  .839}74.57 
		%& \cellcolor[rgb]{ .988,  .894,  .839}136.02 \\
		
		Chg2Cap~\cite{chang2023changes} 
		& \cellcolor[rgb]{ .988,  .894,  .839}86.14 
		& \cellcolor[rgb]{ .988,  .894,  .839}78.08 
		& \cellcolor[rgb]{ .988,  .894,  .839}70.66 
		& \cellcolor[rgb]{ .988,  .894,  .839}64.39 
		& \cellcolor[rgb]{ .988,  .894,  .839}40.03 
		& \cellcolor[rgb]{ .988,  .894,  .839}75.12 
		& \cellcolor[rgb]{ .988,  .894,  .839}136.61 \\
		
		%CVMSL-L~\cite{xian2026cross}
		%& -
		%& -
		%& -
		%& \cellcolor[rgb]{ .988,  .894,  .839}64.83
		%& \cellcolor[rgb]{ 1,  .933,  .859}39.72
		%& \cellcolor[rgb]{ .988,  .894,  .839}74.97
		%& \cellcolor[rgb]{ .988,  .894,  .839}136.59 \\

		MV-CC~\cite{liu2024mv} 
		& \cellcolor[rgb]{ .973,  .796,  .678}86.37 
		& \cellcolor[rgb]{ .973,  .796,  .678}79.01 
		& \cellcolor[rgb]{ .973,  .796,  .678}72.03 
		& \cellcolor[rgb]{ .973,  .796,  .678}66.22 
		& \cellcolor[rgb]{ .988,  .894,  .839}40.20 
		& \cellcolor[rgb]{ .973,  .796,  .678}75.73 
		& \cellcolor[rgb]{ .988,  .894,  .839}138.28 \\
		
		Change3D~\cite{zhu2025change3d} 
		& \cellcolor[rgb]{ .988,  .894,  .839}85.81 
		& \cellcolor[rgb]{ .988,  .894,  .839}77.81 
		& \cellcolor[rgb]{ .988,  .894,  .839}70.57 
		& \cellcolor[rgb]{ .988,  .894,  .839}64.38 
		& \cellcolor[rgb]{ .988,  .894,  .839}40.38 
		& \cellcolor[rgb]{ .988,  .894,  .839}75.12 
		& \cellcolor[rgb]{ .988,  .894,  .839}138.29 \\
		
		KCFI~\cite{yang2025enhancing} 
		& \cellcolor[rgb]{ .973,  .796,  .678}86.34 
		& \cellcolor[rgb]{ .988,  .894,  .839}77.31 
		& \cellcolor[rgb]{ .988,  .894,  .839}70.89 
		& \cellcolor[rgb]{ .988,  .894,  .839}65.30 
		& \cellcolor[rgb]{ 1,  .933,  .859}39.42 
		& \cellcolor[rgb]{ .988,  .894,  .839}75.47 
		& \cellcolor[rgb]{ .988,  .894,  .839}138.25 \\
		
		% CTSD-Net~\cite{wu2025cross} 
		% & \cellcolor[rgb]{ .973,  .796,  .678}86.75 
		% & \cellcolor[rgb]{ .973,  .796,  .678}78.94 
		% & \cellcolor[rgb]{ .973,  .796,  .678}72.06 
		% & \cellcolor[rgb]{ .973,  .796,  .678}66.32 
		% & \cellcolor[rgb]{ .973,  .796,  .678}40.34 
		% & \cellcolor[rgb]{ .973,  .796,  .678}75.70 
		% & \cellcolor[rgb]{ .973,  .796,  .678}140.22 \\
		
		CD4C~\cite{li2025cd4c} 
		& \cellcolor[rgb]{ .973,  .796,  .678}86.74 
		& \cellcolor[rgb]{ .973,  .796,  .678}78.71 
		& \cellcolor[rgb]{ .973,  .796,  .678}71.48 
		& \cellcolor[rgb]{ .973,  .796,  .678}65.41 
		& \cellcolor[rgb]{ .973,  .796,  .678}40.56 
		& \cellcolor[rgb]{ .973,  .796,  .678}75.72 
		& \cellcolor[rgb]{ .988,  .894,  .839}138.00 \\
		
		SCNet~\cite{sun2026scnet}
		& \cellcolor[rgb]{ .973,  .796,  .678}86.14
		& \cellcolor[rgb]{ .988,  .894,  .839}78.19
		& \cellcolor[rgb]{ .973,  .796,  .678}71.44
		& \cellcolor[rgb]{ .973,  .796,  .678}65.82
		& \cellcolor[rgb]{ .973,  .796,  .678}40.51
		& \cellcolor[rgb]{ .988,  .894,  .839}75.37
		& \cellcolor[rgb]{ .973,  .796,  .678}140.23 \\
		
		RMNet~\cite{cao2026rmnet}
		& \cellcolor[rgb]{ .988,  .894,  .839}86.09
		& \cellcolor[rgb]{ .988,  .894,  .839}78.43
		& \cellcolor[rgb]{ .973,  .796,  .678}71.68
		& \cellcolor[rgb]{ .973,  .796,  .678}66.11
		& \cellcolor[rgb]{ .973,  .796,  .678}40.94
		& \cellcolor[rgb]{ .973,  .796,  .678}75.57
		& \cellcolor[rgb]{ .973,  .796,  .678}139.43 \\
		
		D3-Net~\cite{peng2026frequency}
		& \cellcolor[rgb]{ .973,  .796,  .678}86.70
		& \cellcolor[rgb]{ .973,  .796,  .678}78.77
		& \cellcolor[rgb]{ .973,  .796,  .678}71.60
		& \cellcolor[rgb]{ .973,  .796,  .678}65.68
		& \cellcolor[rgb]{ .973,  .796,  .678}40.74
		& \cellcolor[rgb]{ .973,  .796,  .678}75.82
		& \cellcolor[rgb]{ .973,  .796,  .678}140.52 \\

		%DualFocus-CapNet~\cite{meng2025dualfocus}
		%& \cellcolor[rgb]{ .988,  .894,  .839}85.47
		%& \cellcolor[rgb]{ .988,  .894,  .839}77.67
		%& \cellcolor[rgb]{ .973,  .796,  .678}71.19
		%& \cellcolor[rgb]{ .973,  .796,  .678}66.00
		%& \cellcolor[rgb]{ .973,  .796,  .678}40.78
		%& \cellcolor[rgb]{ .973,  .796,  .678}75.51
		%& \cellcolor[rgb]{ .973,  .796,  .678}139.36 \\

		CTM~\cite{bai2025cross} 
		& \cellcolor[rgb]{ .973,  .796,  .678}86.78 
		& \cellcolor[rgb]{ .973,  .796,  .678}79.24 
		& \cellcolor[rgb]{ .973,  .796,  .678}72.65 
		& \cellcolor[rgb]{ .973,  .796,  .678}65.48 
		& \cellcolor[rgb]{ .973,  .796,  .678}40.63 
		& \cellcolor[rgb]{ .973,  .796,  .678}75.72 
		& \cellcolor[rgb]{ .973,  .796,  .678}138.78 \\
		
		DGAT~\cite{qin2025dgat}
		& \cellcolor[rgb]{ .973,  .796,  .678}87.09
		& \cellcolor[rgb]{ .973,  .796,  .678}78.93
		& \cellcolor[rgb]{ .973,  .796,  .678}71.35
		& \cellcolor[rgb]{ .973,  .796,  .678}65.30
		& \cellcolor[rgb]{ .973,  .796,  .678}40.98
		& \cellcolor[rgb]{ .973,  .796,  .678}76.41
		& \cellcolor[rgb]{ .973,  .796,  .678}141.27 \\

		\hline
		\textbf{STAND (ours)} 
		& \cellcolor[rgb]{ .957,  .69,  .518}\textbf{88.24} 
		& \cellcolor[rgb]{ .957,  .69,  .518}\textbf{80.33} 
		& \cellcolor[rgb]{ .957,  .69,  .518}\textbf{73.15} 
		& \cellcolor[rgb]{ .957,  .69,  .518}\textbf{67.11} 
		& \cellcolor[rgb]{ .957,  .69,  .518}\textbf{41.55} 
		& \cellcolor[rgb]{ .957,  .69,  .518}\textbf{77.25} 
		& \cellcolor[rgb]{ .957,  .69,  .518}\textbf{143.39} \\
		\hline
	\end{tabular}
	\caption{Comparison with other methods on LEVIR-CC.}
	\label{tab:comparison_levir}
\end{table*}

\begin{table*}[t]
	\centering
	\small
	\renewcommand{\arraystretch}{1.15}
	\begin{tabular}{l|ccccccc}
		\hline
		\rule{0pt}{12pt}\footnotesize Method & \footnotesize BLEU-1 & \footnotesize BLEU-2 & \footnotesize BLEU-3 & \footnotesize BLEU-4 & \footnotesize METEOR & \footnotesize ROUGE$_{L}$ & \footnotesize CIDEr \\
		\hline
		%PromptCC~\cite{liu2023decoupling} 
		%& \cellcolor[rgb]{ 1,  .933,  .859}81.12
		%& \cellcolor[rgb]{ 1,  .933,  .859}73.96
		%& \cellcolor[rgb]{ 1,  .933,  .859}67.22
		%& \cellcolor[rgb]{ 1,  .933,  .859}61.45
		%& 36.99
		%& \cellcolor[rgb]{ 1,  .933,  .859}71.88
		%& 134.50 \\
		
		RSICCFormer~\cite{liu2022remote} 
		& \cellcolor[rgb]{ 1,  .933,  .859}80.05
		& \cellcolor[rgb]{ 1,  .933,  .859}74.24
		& \cellcolor[rgb]{ 1,  .933,  .859}69.61
		& \cellcolor[rgb]{ 1,  .933,  .859}66.54
		& \cellcolor[rgb]{ 1,  .933,  .859}42.65
		& \cellcolor[rgb]{ 1,  .933,  .859}73.91
		& 133.44 \\
		
		CTMTNet~\cite{shi2024multi} 
		& \cellcolor[rgb]{ 1,  .933,  .859}83.56
		& \cellcolor[rgb]{ .988,  .894,  .839}77.66 
		& \cellcolor[rgb]{ .988,  .894,  .839}72.76
		& \cellcolor[rgb]{ .988,  .894,  .839}69.00
		& \cellcolor[rgb]{ .988,  .894,  .839}45.39
		& \cellcolor[rgb]{ .988,  .894,  .839}79.23
		& \cellcolor[rgb]{ 1,  .933,  .859}149.40 \\
		
		CTM~\cite{bai2025cross} 
		& \cellcolor[rgb]{ .988,  .894,  .839}85.36
		& \cellcolor[rgb]{ .988,  .894,  .839}79.49
		& \cellcolor[rgb]{ .988,  .894,  .839}75.36
		& \cellcolor[rgb]{ .988,  .894,  .839}72.36
		& \cellcolor[rgb]{ .988,  .894,  .839}46.98
		& \cellcolor[rgb]{ .988,  .894,  .839}80.97
		& \cellcolor[rgb]{ .988,  .894,  .839}153.29 \\
		
		%Chg2Cap~\cite{chang2023changes} 
		%& \cellcolor[rgb]{ .988,  .894,  .839}86.33
		%& \cellcolor[rgb]{ .973,  .796,  .678}82.07
		%& \cellcolor[rgb]{ .973,  .796,  .678}78.54
		%& \cellcolor[rgb]{ .973,  .796,  .678}76.16 
		%& \cellcolor[rgb]{ .973,  .796,  .678}48.29
		%& \cellcolor[rgb]{ .973,  .796,  .678}81.44
		%& \cellcolor[rgb]{ .988,  .894,  .839}156.10 \\
		
		SEN~\cite{att1zhou2024single} 
		& \cellcolor[rgb]{ .988,  .894,  .839}85.12 
		& \cellcolor[rgb]{ .988,  .894,  .839}80.28
		& \cellcolor[rgb]{ .973,  .796,  .678}76.84
		& \cellcolor[rgb]{ .973,  .796,  .678}74.71
		& \cellcolor[rgb]{ .988,  .894,  .839}46.81 
		& \cellcolor[rgb]{ .988,  .894,  .839}79.45 
		& \cellcolor[rgb]{ .988,  .894,  .839}148.70 \\
		
		CTSD-Net~\cite{wu2025cross} 
		& \cellcolor[rgb]{ .988,  .894,  .839}85.49
		& \cellcolor[rgb]{ .988,  .894,  .839}80.07
		& \cellcolor[rgb]{ .973,  .796,  .678}76.45
		& \cellcolor[rgb]{ .973,  .796,  .678}73.84
		& \cellcolor[rgb]{ .973,  .796,  .678}47.75
		& \cellcolor[rgb]{ .988,  .894,  .839}80.99 
		& \cellcolor[rgb]{ .988,  .894,  .839}153.96 \\
		
		D3-Net~\cite{peng2026frequency}
		& \cellcolor[rgb]{ .988,  .894,  .839}85.42
		& \cellcolor[rgb]{ .988,  .894,  .839}80.02
		& \cellcolor[rgb]{ .988,  .894,  .839}76.23
		& \cellcolor[rgb]{ .988,  .894,  .839}73.15
		& \cellcolor[rgb]{ .988,  .894,  .839}47.13
		& \cellcolor[rgb]{ .973,  .796,  .678}81.47
		& \cellcolor[rgb]{ .988,  .894,  .839}154.33 \\

		DualFocus-CapNet~\cite{meng2025dualfocus}
		& \cellcolor[rgb]{ .988,  .894,  .839}85.85
		& \cellcolor[rgb]{ .988,  .894,  .839}80.67
		& \cellcolor[rgb]{ .973,  .796,  .678}76.65
		& \cellcolor[rgb]{ .973,  .796,  .678}73.59
		& \cellcolor[rgb]{ .973,  .796,  .678}48.03
		& \cellcolor[rgb]{ .973,  .796,  .678}81.28
		& \cellcolor[rgb]{ .973,  .796,  .678}157.71 \\

		FST-Net~\cite{zou2025frequency} 
		& \cellcolor[rgb]{ .973,  .796,  .678}88.15
		& \cellcolor[rgb]{ .973,  .796,  .678}83.46
		& \cellcolor[rgb]{ .973,  .796,  .678}79.55
		& \cellcolor[rgb]{ .957,  .69,  .518}76.78
		& \cellcolor[rgb]{ .973,  .796,  .678}48.78
		& \cellcolor[rgb]{ .973,  .796,  .678}82.91
		& \cellcolor[rgb]{ .973,  .796,  .678}160.01 \\
		
		\hline
		\textbf{STAND (ours)} 
		& \cellcolor[rgb]{ .957,  .69,  .518}\textbf{88.66} 
		& \cellcolor[rgb]{ .957,  .69,  .518}\textbf{84.04} 
		& \cellcolor[rgb]{ .957,  .69,  .518}\textbf{80.05} 
		& \cellcolor[rgb]{ .957,  .69,  .518}\textbf{76.91} 
		& \cellcolor[rgb]{ .957,  .69,  .518}\textbf{49.08} 
		& \cellcolor[rgb]{ .957,  .69,  .518}\textbf{83.36} 
		& \cellcolor[rgb]{ .957,  .69,  .518}\textbf{160.78} \\
		\hline
	\end{tabular}
	\caption{Comparison with other methods on WHU-CDC.}
	\label{tab:comparison_whucdc}
\vspace{-3pt}
\end{table*}

\begin{table*}[t]
	\centering
	\footnotesize
	\renewcommand{\arraystretch}{1.15}
	\begin{tabular*}{0.76\textwidth}{@{\extracolsep{\fill}}clccccc@{}}
		\multicolumn{7}{c}{\textit{(a) Progressive component ablation}} \\
		\toprule
		\# & Model & BLEU-4 & METEOR & ROUGE$_L$ & CIDEr & $\Delta$CIDEr \\
		\midrule
		B0 & Video Encoder (base)        & 64.38 & 40.38 & 75.12 & 138.29 & -- \\
		B1 & \!+\! DGTD (CAVD+FRCA)          & 65.24 & 40.31 & 75.81 & 139.29 & +1.00 \\
		B2 & \!+\! DGTD (CAVD+FRCA) \!+\! SCA    & 65.82 & 41.03 & 76.78 & 142.11 & +3.82 \\
		\rowcolor{gray!15}
		B3 & + ITC + DGTD + SCA (Full)   & \textbf{67.11} & \textbf{41.55} & \textbf{77.25} & \textbf{143.39} & +5.10 \\
		\bottomrule
	\end{tabular*}

	\vspace{4pt}

	\begin{tabular*}{0.76\textwidth}{@{\extracolsep{\fill}}lcccccc@{}}
		\multicolumn{7}{c}{\textit{(b) DGTD component ablation}} \\
		\toprule
		Setting & CAVD & FRCA & BLEU-4 & METEOR & ROUGE$_L$ & CIDEr \\
		\midrule
		w/o DGTD          & $\times$ & $\times$ & 64.38 & 40.25 & 75.61 & 139.66 \\
		Macro only (CAVD) & $\checkmark$ & $\times$ & \textbf{65.52} & 40.41 & 75.83 & 140.21 \\
		Micro only (FRCA) & $\times$ & $\checkmark$ & 64.79 & 40.55 & 75.58 & 140.17 \\
		\rowcolor{gray!15}
		Dual (CAVD+FRCA)  & $\checkmark$ & $\checkmark$ & 65.47 & \textbf{40.60} & \textbf{76.39} & \textbf{141.22} \\
		\bottomrule
	\end{tabular*}
	\caption{Ablation studies on LEVIR-CC: (a) progressive component ablation; (b) DGTD ablation with ITC enabled and SCA excluded.}
	\label{tab:ablation_main}
\end{table*}

\section{Experiments}
\label{sec:Experiments}

\subsection{Experimental Setup}

\textbf{Datasets.}
LEVIR-CC~\cite{liu2022remote} consists of 10{,}077 image pairs with 5{,}038 change pairs.
Each pair is annotated with five sentences. 
% We follow the official split with 6{,}815 pairs for training, 1{,}333 for validation, and 
% 1{,}929 for testing.
WHU-CDC~\cite{shi2024multi} comprises 7{,}434 image pairs and 37{,}170 corresponding 
change descriptions. 
The sentences range from 3 to 24 words and are drawn from a vocabulary of 327 tokens. 
Both datasets adopt the same five annotated object categories, including \emph{building}, \emph{road}, \emph{parking lot}, \emph{vegetation}, and \emph{water}.

\noindent\textbf{Evaluation metrics.}
Following prior work~\cite{liu2022remote}, we evaluate the captions with seven common metrics, 
\textit{i.e.}, BLEU-N ($N=1,2,3,4$)~\cite{papineni2002bleu}, 
METEOR~\cite{banerjee2005meteor}, 
ROUGE$_L$~\cite{lin2004rouge}, and CIDEr-D~\cite{vedantam2015cider}.

\noindent\textbf{Implementation details.}
The proposed method adopts X3D-L~\cite{feichtenhofer2020x3d} as the baseline. 
All experiments are conducted on an RTX~4090 GPU. 
% After encoding, each frame feature has a spatial size of $192 \times 16 \times 16$.
%To reduce GPU memory consumption, we employ mixed-precision training. 
% The batch size is set to 16, and the word embedding dimension is 192. 
% We train the model for up to 20 epochs, with a beam size of 1 during inference.
% In addition, we call the ChatGPT API to extract object category change labels from the dataset captions, and the change masks are obtained from pre-trained binary change detection models. 
To extract category change labels for the training set, we construct a dual-agent verification: Agent A (Qwen2.5-72B-Instruct) derives labels from only the training set, while Agent B (GPT-4o) performs an inspection.
In addition, the change masks are obtained from a pre-trained change detector. 
\textit{\textbf{Importantly, label extraction and detector pretraining procedures are conducted strictly within the training data, thereby avoiding data leakage.}}

\subsection{Performance Comparison}

\subsubsection{Results on the LEVIR-CC Dataset.}
We compare STAND with SOTA methods, including Chg2Cap~\cite{chang2023changes}, MV-CC~\cite{liu2024mv}, CD4C~\cite{li2025cd4c}, Change3D~\cite{zhu2025change3d}, CTM~\cite{bai2025cross}, KCFI~\cite{yang2025enhancing}, SCNet~\cite{sun2026scnet},  RMNet~\cite{cao2026rmnet}, D3-Net~\cite{peng2026frequency}, and DGAT~\cite{qin2025dgat}.

As shown in~\cref{tab:comparison_levir}, STAND achieves the best performance , 
with notable improvements in METEOR (+0.57), ROUGE$_L$ (+0.84), and CIDEr (+2.12) over the second-best method (DGAT). 
Meanwhile, MV-CC~\cite{liu2024mv} and CD4C~\cite{li2025cd4c} are the methods that apply masks via simple filtering, whereas STAND leverages the mask as a guidance to focus on changes. 
STAND outperforms both approaches indicating the superiority of this guidance method.

\subsubsection{Results on the WHU-CDC Dataset.}
We compare STAND with state-of-the-art methods, including  RSICCFormer~\cite{liu2022remote}, CTMTNet~\cite{shi2024multi},
CTM~\cite{bai2025cross}, SEN~\cite{att1zhou2024single}, CTSD-Net~\cite{wu2025cross}, FST-Net~\cite{zou2025frequency}, D3-Net~\cite{peng2026frequency}, and DualFocus-CapNet~\cite{meng2025dualfocus}.

% As shown in~\cref{tab:comparison_whucdc}, STAND still surpasses all methods, 
% achieving a CIDEr score of 160.78, outperforming the second-best method FST-Net by a significant margin. 
% Besides, we note that SEN adopts a single-stream architecture, but it still treats the images as a static pair and models change at the difference-feature level, which limits its ability to reason about the change. This further demonstrates the superiority of dynamic spatio-temporal modeling.

Existing methods already achieve high scores on WHU-CDC, leaving limited room for further improvement. Nevertheless, STAND obtains the highest reported scores across all seven metrics in~\cref{tab:comparison_whucdc}, with modest but consistent gains over FST-Net and CIDEr improvements of 6.45 and 3.07 points over the recent D3-Net and DualFocus-CapNet, respectively.

\subsection{Ablation Study and Analysis}

% \vspace{-20pt}

% \begin{table}[t]
% \centering
% \footnotesize
% \setlength{\tabcolsep}{2pt}
% \renewcommand{\arraystretch}{1.15}
% \caption{Ablation of model components on LEVIR-CC.
% S0 denotes the full \textbf{STAND} model with all modules enabled.}
% \label{tab:ablation_modules}

% \begin{tabular}{lcccccccc}
% \toprule
% Setting
% & ITC
% & CAVD 
% & FRCA 
% & SCA 
% & BLEU-4 
% & METEOR 
% & ROUGE$_L$ 
% & CIDEr \\
% \midrule

% \rowcolor{gray!15}
% \textbf{S0 (Full)}
% & $\checkmark$ & $\checkmark$ & $\checkmark$ & $\checkmark$
% & \textbf{67.11}
% & \textbf{41.55}
% & \textbf{77.25}
% & \textbf{143.39} \\

% S1 & $\checkmark$ & $\checkmark$ & $\checkmark$ & $\times$
% & 65.15 & 40.57 & 76.42 & 141.20 \\

% S2 & $\checkmark$ & $\checkmark$ & $\times$ & $\times$
% & 65.54 & 40.11 & 75.84 & 137.27 \\

% S3 & $\checkmark$ & $\times$ & $\checkmark$ & $\times$
% & 64.78 & 40.15 & 75.58 & 138.67 \\

% S4 & $\times$ & $\checkmark$ & $\checkmark$ & $\times$
% & 64.24 & 40.31 & 75.81 & 139.26 \\

% S5 & $\times$ & $\checkmark$ & $\checkmark$ & $\checkmark$
% & 65.83 & 41.03 & 76.79 & 143.21 \\

% S6 & $\checkmark$ & $\times$ & $\checkmark$ & $\checkmark$
% & 65.51 & 40.42 & 75.61 & 138.91 \\

% S7 & $\checkmark$ & $\checkmark$ & $\times$ & $\checkmark$
% & 65.58 & 40.83 & 76.02 & 140.69 \\

% S8 & $\checkmark$ & $\times$ & $\times$ & $\checkmark$
% & 65.38 & 40.25 & 75.61 & 140.26 \\

% \bottomrule
% \end{tabular}
% \end{table}

\noindent\textbf{Progressive Ablation on LEVIR-CC.}
As shown in~\cref{tab:ablation_main}(a),
the model performance steadily improves as modules are progressively added. 
Among them, incorporating the SCA module brings the most significant performance gain, particularly in CIDEr. 
Adding ITC leads to improvements in absolute scores, and we conduct five other independent runs on B2 \& B3 which demonstrate it stabilizes training, 
reducing the averaged metric variance from \underline{\textit{0.26}} to \underline{\textit{0.18}} across all evaluation metrics. 
These results verify the effectiveness and stability of the proposed progressive refinement pipeline.

\noindent\textbf{Analysis of DGTD.}
Viewpoint and scale ambiguities are inherently spatial in nature, fundamentally different from knowledge ambiguity. 
To isolate their effects, we evaluate CAVD and FRCA separately in~\cref{tab:ablation_main}(b), while keeping ITC enabled and excluding SCA for clarity.
Using either CAVD or FRCA alone brings limited gains, indicating that resolving spatial ambiguity from a single perspective is insufficient. 
In contrast, combining both components yields the best performance on most metrics, demonstrating that macro-level contextual conditioning and micro-level feature refinement are complementary for tackling spatial ambiguity.

% Overall, these results demonstrate that the proposed modules contribute at distinct stages of representation refinement, jointly leading to more precise and semantically consistent change captions.

\noindent\textbf{Effect of the Encoder Architecture.}
We further investigate image- and video-based backbones, including ResNet-101~\cite{he2016deep}, I3D~\cite{video3}, Swin-T~\cite{liu2022video}, and X3D~\cite{feichtenhofer2020x3d}.
Among them, X3D-L achieves the best overall balance across BLEU, METEOR, and ROUGE.
Considering its stable and competitive performance across metrics, we adopt X3D-L as the default backbone in subsequent experiments.

\begin{table}[t]
\centering
\footnotesize
\setlength{\tabcolsep}{2pt}
\renewcommand{\arraystretch}{1.0}

\begin{tabular}{@{}lcccc@{}}
\toprule
Setting
& BLEU-4
& METEOR 
& ROUGE$_L$
& CIDEr \\

\midrule
\multicolumn{5}{@{}l}{\textit{Encoder Architecture}} \\

\rowcolor{backbone!50}
ResNet-101 (Image)
& 65.13 & 40.45 & 75.50 & 139.81 \\

\rowcolor{backbone!50}
I3D
& \cellcolor{backbone}66.71 & 41.00 & 76.02 & 142.36 \\

\rowcolor{backbone!50}
Swin-T
& \cellcolor{backbone}67.03 & \cellcolor{backbone}41.23 & \cellcolor{backbone}76.98 & \cellcolor{backbone}\textbf{143.66} \\

% \rowcolor{backbone!50}
% X3D-XS  
% & 64.74 & 39.90 & 74.95 & 137.46 \\

% \rowcolor{backbone!50}
% X3D-S   
% & 64.28 & 40.44 & 75.28 & 138.00 \\

\rowcolor{backbone!50}
X3D-M   
& 64.52 & 40.21 & 75.42 & 138.53 \\

\rowcolor{backbone!50}
X3D-L   
& \cellcolor{backbone}\textbf{67.11}
& \cellcolor{backbone}\textbf{41.55}
& \cellcolor{backbone}\textbf{77.25}
& \cellcolor{backbone}143.39 \\

\midrule
\multicolumn{5}{@{}l}{\textit{Change guidance}} \\

\rowcolor{guidance!50}
Naive Diff. Mask
& \cellcolor{guidance}66.15 & 41.02 & 75.84 & 142.01 \\

\rowcolor{guidance!50}
Pre-trained Mask
& \cellcolor{guidance}\textbf{67.11}
& \cellcolor{guidance}\textbf{41.55}
& \cellcolor{guidance}\textbf{77.25}
& \cellcolor{guidance}\textbf{143.39} \\

% \midrule
% \multicolumn{5}{@{}l}{\textit{Number of Negative Compositions ($n+1$)}} \\

% \rowcolor{negative!50}
% 4
% & 65.70 & 41.00 & \cellcolor{negative}76.47 & \cellcolor{negative}142.20 \\

% \rowcolor{negative!50}
% 3
% & 65.67 & 41.01 & 76.04 & 140.87 \\

% \rowcolor{negative!50}
% 2
% & 65.55 & 40.99 & \cellcolor{negative}76.20 & 141.15 \\

% \rowcolor{negative!50}
% 1
% & \cellcolor{negative}\textbf{67.11}
% & \cellcolor{negative}\textbf{41.55}
% & \cellcolor{negative}\textbf{77.25}
% & \cellcolor{negative}\textbf{143.39} \\

% \midrule

% \textit{Fusion Strategies} \\

% \rowcolor{fusion!50}
% Concat-Conv  
% & 64.63 & 40.32 & 75.07 & 138.94 \\

% \rowcolor{fusion!50}
% iAFF~\cite{dai2021attentional}
% & 65.75 & 40.95 & \cellcolor{fusion}76.31 & 140.71 \\

% \rowcolor{fusion!50}
% Add  
% & 65.92 & \cellcolor{fusion}41.11 & \cellcolor{fusion}76.53 & 142.55 \\

% \rowcolor{fusion!50}
% AFF~\cite{dai2021attentional}
% & \cellcolor{fusion}\textbf{67.11}
% & \cellcolor{fusion}\textbf{41.55}
% & \cellcolor{fusion}\textbf{77.25}
% & \cellcolor{fusion}\textbf{143.39} \\

\bottomrule
\end{tabular}
\caption{Ablation studies on LEVIR-CC: encoder and masks.}
\label{tab:combined_ablations}

\end{table}

\begin{figure*}[t]
  \centering
   %\setlength{\abovecaptionskip}{-10cm}
  % \fbox{\rule{0pt}{0.5in} \rule{0.9\linewidth}{0pt}}
  \includegraphics[width=0.85\linewidth]{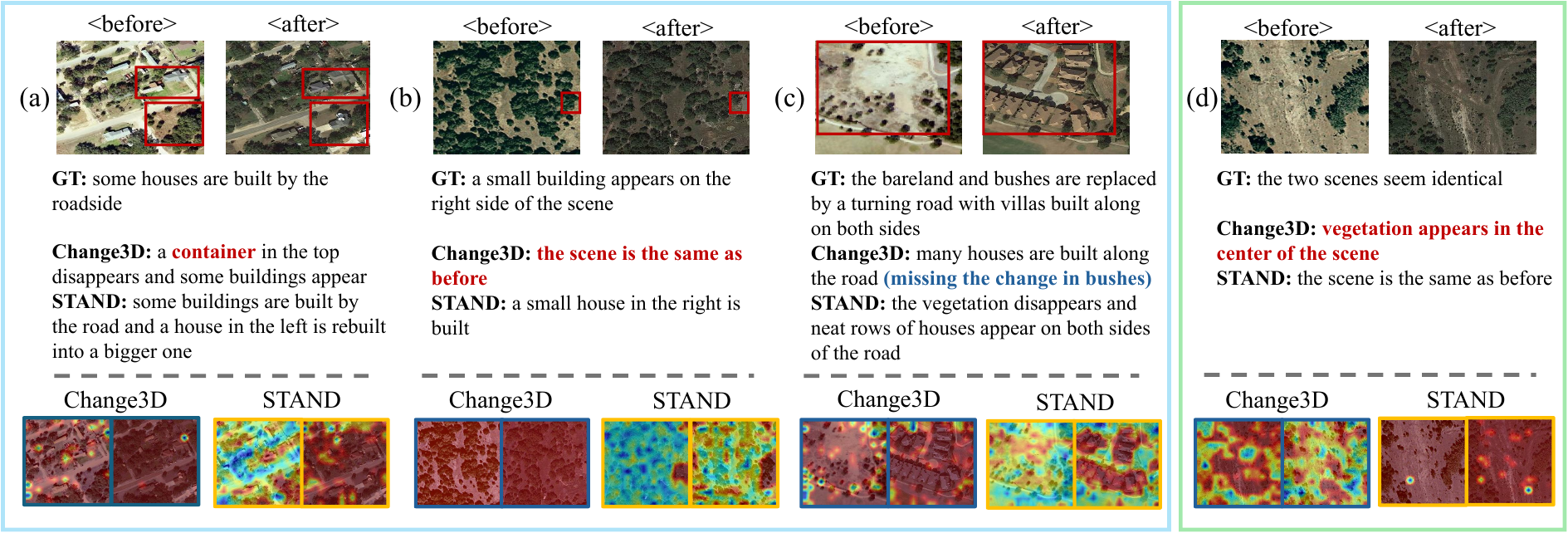}
   \caption{Visualization of baseline Change3D~\cite{zhu2025change3d} and STAND on LEVIR-CC.  (a), (b), and (c) are change image pairs, while (d) is a no-change pair.}
   \label{heat_map}
\end{figure*}

\begin{table*}[t]
\centering
\footnotesize
\setlength{\tabcolsep}{3pt}
\renewcommand{\arraystretch}{0.85}

\begin{tabular}[t]{lcc}
\multicolumn{3}{c}{\textit{(a) Scale (Small-Scale Subset)}} \\
\toprule
Setting & BLEU-4 & METEOR \\
\midrule
Change3D (Baseline) & 30.06 & 20.12 \\
\rowcolor{gray!8}
Ours w/o FRCA & 30.72 & 20.86 \\
\rowcolor{gray!15}
Ours & \cellcolor{gray!30}\textbf{31.39} & \cellcolor{gray!30}\textbf{21.91} \\
\bottomrule
\end{tabular}
\hfill
\begin{tabular}[t]{lcc}
\multicolumn{3}{c}{\textit{(b) Viewpoint Ambiguity}} \\
\toprule
Setting & FPR$\downarrow$ & Prec$\uparrow$ \\
\midrule
Change3D (Baseline) & 0.06 & 0.70 \\
\rowcolor{gray!8}
Ours w/o CAVD & 0.04 & 0.74 \\
\rowcolor{gray!15}
Ours & \cellcolor{gray!30}\textbf{0.02} & \cellcolor{gray!30}\textbf{0.78} \\
\bottomrule
\end{tabular}
\hfill
\begin{tabular}[t]{lccc}
\multicolumn{4}{c}{\textit{(c) Knowledge Ambiguity}} \\
\toprule
Setting & Prec$\uparrow$ & Rec$\uparrow$ & F1$\uparrow$ \\
\midrule
Change3D (Baseline) & 0.70 & 0.72 & 0.71 \\
\rowcolor{gray!8}
Ours w/o SCA & 0.73 & 0.77 & 0.75 \\
\rowcolor{gray!15}
Ours & \cellcolor{gray!30}\textbf{0.78} & \cellcolor{gray!30}\textbf{0.85} & \cellcolor{gray!30}\textbf{0.81} \\
\bottomrule
\end{tabular}
\caption{Ambiguity-specific evaluation on LEVIR-CC.}
\label{tab:ambiguity_analysis}
\vspace{-3pt}
\end{table*}

\noindent\textbf{Effect of the Change Mask.}
To verify the necessity of high-quality change guidance, we replace the change mask with a naive pixel-wise difference mask. Specifically, this mask is generated by computing the absolute pixel difference between two images and applying a binary threshold. As shown in~\cref{tab:combined_ablations},  while this replacement leads to a metric drop, our method equipped with this naive mask \textit{still} outperforms all \textit{non-mask-based} models in~\cref{tab:comparison_levir}. This demonstrates that the superiority of STAND does not merely rely on the external mask, but fundamentally stems from the disambiguation capabilities of our core pipeline.

\noindent\textbf{Ambiguity-Specific Evaluation.}
Existing RSICC works lack targeted evaluation for ambiguity.
To validate whether STAND mitigates ambiguities, 
we design metrics for different ambiguity types, as shown in~\cref{tab:ambiguity_analysis}. 

For \textit{scale ambiguity}, we evaluate results on a small-scale subset (selected by LLM), 
where BLEU-4 and METEOR reflect the model's ability to describe subtle changes. 
For \textit{viewpoint ambiguity}, we report the false positive rate (FPR) and precision, which measure whether the model avoids hallucinated changes when entities share similar appearance.
For \textit{knowledge ambiguity}, we further report entity-level recall and F1 score, which quantify the completeness and balance of detected changed categories.
All entity-level metrics (FPR, Prec, Rec, F1) are computed by extracting change labels from the generated captions and comparing them with ground-truth labels.
These are standard evaluation metrics in captioning and machine learning.
Compared with Change3D and the targeted ablations, the full model performs best, confirming the roles of FRCA, CAVD, and SCA in scale, viewpoint, and knowledge disambiguation, respectively.
% \noindent\textbf{Effect of Different Fusion Strategies.} 
% We investigate how different fusion strategies in~\cref{eq:fusion} affect the transition constraint process. 
% Specifically, we compare four strategies: (i) concatenation followed by convolution (Concat-Conv), (ii) element-wise addition (Add), (iii) iAFF~\cite{dai2021attentional}, and (iv) AFF~\cite{dai2021attentional}. 
% As reported in~\cref{tab:combined_ablations}, AFF achieves the best results, showing its effectiveness in the fusion.

\begin{center}
	\begin{minipage}{0.95\linewidth}
		\centering
		\includegraphics[width=\linewidth]{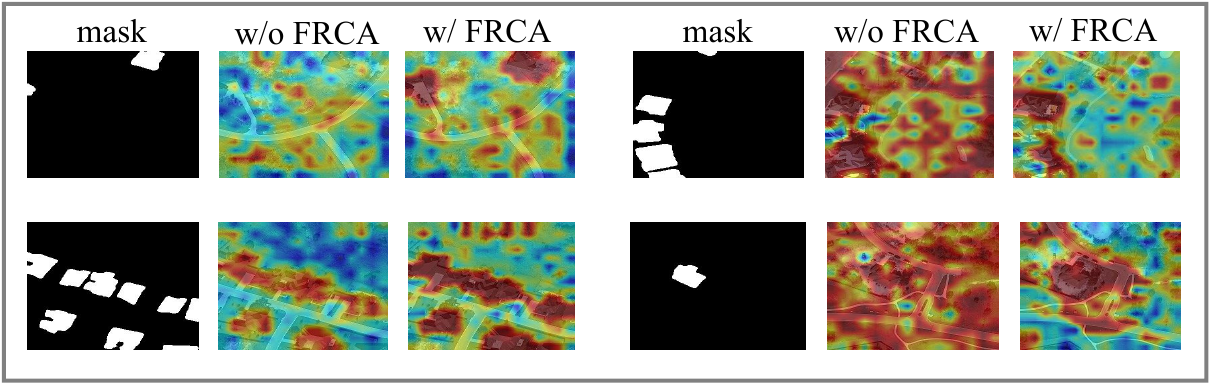}
		\captionof{figure}{Visualization of the heatmap, without (left) and with (right) FRCA.}
		\label{att}
	\end{minipage}
\end{center}

\subsection{Visualization} \label{sec:vis}
We further compare STAND with Change3D~\cite{zhu2025change3d} and present four representative cases.
As shown in \cref{heat_map}, we can observe: 
(a) Under the top-down remote-sensing viewpoint, the baseline misreads the building as a container, whereas our approach correctly recognizes the change. 
(b) When changes occur at a small scale, our model reliably captures differences, whereas Change3D predicts no change.
(c) For captions involving multiple objects, STAND effectively exploits prior knowledge to improve semantic precision. In contrast, the caption generated by Change3D misses some of the changed objects.
(d) For image pairs with no actual change, our attention remains uniformly distributed and is robust to seasonal effects. 
In addition, as shown in \cref{att}, the proposed FRCA module refocuses attention from low-frequency regions to true change areas, improving the localization of real changes.

\begin{table}[t]
	\centering
	\footnotesize
	\setlength{\tabcolsep}{1pt}
	\renewcommand{\arraystretch}{1.1}
	\begin{tabular}{@{}lc|cccc@{}}
		\hline
		\rule{0pt}{12pt}\footnotesize Method & \footnotesize Model Size & \footnotesize BLEU-4 & \footnotesize METEOR & \footnotesize ROUGE$_{L}$ & \footnotesize CIDEr \\
		\hline
		MiniGPT4 & 1.1B + 7B
		& \cellcolor[rgb]{0.85,0.93,1.0}62.55 
		& \cellcolor[rgb]{0.85,0.93,1.0}39.60 
		& \cellcolor[rgb]{0.85,0.93,1.0}74.22 
		& \cellcolor[rgb]{0.85,0.93,1.0}136.10 \\
		
		LLaVA-1.5 & 0.3B + 7B
		& \cellcolor[rgb]{0.85,0.93,1.0}62.75
		& \cellcolor[rgb]{0.85,0.93,1.0}39.85
		& \cellcolor[rgb]{0.85,0.93,1.0}75.10
		& \cellcolor[rgb]{0.85,0.93,1.0}136.56 \\
		
		LLaVA-ov & 0.4B + 0.5B
		& \cellcolor[rgb]{0.85,0.93,1.0}64.55
		& \cellcolor[rgb]{0.85,0.93,1.0}40.77
		& \cellcolor[rgb]{0.65,0.82,1.0}75.32
		& \cellcolor[rgb]{0.85,0.93,1.0}138.24 \\
		
		LLaVA-ov & 0.4B + 7B
		& \cellcolor[rgb]{0.65,0.82,1.0}63.66
		& \cellcolor[rgb]{0.65,0.82,1.0}41.23
		& \cellcolor[rgb]{0.65,0.82,1.0}75.78
		& \cellcolor[rgb]{0.65,0.82,1.0}139.61 \\
		
		STAND & 9.2M + 6.5M 
		& \cellcolor[rgb]{0.45,0.70,1.0}67.11
		& \cellcolor[rgb]{0.45,0.70,1.0}41.55
		& \cellcolor[rgb]{0.45,0.70,1.0}77.25
		& \cellcolor[rgb]{0.45,0.70,1.0}143.39 \\
		\hline
	\end{tabular}
	\caption{Comparison with MLLM methods on LEVIR-CC. Model size represents encoder + decoder.}
	\label{tab:mllm}
\vspace{-10pt}
\end{table}

\subsection{Comparison with MLLM Methods}
To further examine whether general MLLMs can replace task-specific architectures and achieve competitive performance, we adapt off-the-shelf MLLMs~\cite{zhu2023minigpt,liu2023visual,li2024llava} to RS change captioning and perform instruction tuning on the training split.
As shown in~\cref{tab:mllm}, our STAND clearly surpasses the tuned MLLMs. 
This suggests that MLLMs still lack sufficient domain awareness for this vertical remote sensing task, making their generic priors difficult to transfer to RSICC.
Therefore, in the era of foundation MLLMs, carefully designed compact models still remain indispensable for RSICC, especially when domain-specific (e.g., remote sensing) data for pretraining or fine-tuning are limited in practice.

\section{Conclusion}
In this paper, we proposed STAND to model changes in a spatio-temporal manner, thereby resolving the ambiguities of viewpoint, scale, and knowledge in RS images. 
In the future, we plan to further explore the applications of RSICC to high-resolution images and more complex scenes (e.g., deserts, rainforests, coastal zones).
At the same time, we hope to develop a closer integration with MLLMs and remote sensing priors to generate more high-level semantics.
We believe that our work has the potential to inspire more studies on RSICC with spatio-temporal modeling. 

\bibliography{main}
\end{document}